\documentclass[5p,final,authoryear,twocolumn]{elsarticle}

\usepackage{amssymb}
\usepackage{graphicx}
\usepackage{amsmath}
\usepackage{subfigure}
\usepackage[latin1]{inputenc}
\usepackage{color}
\usepackage{multirow}
\usepackage[hypcap=false]{caption}
\usepackage{rotating}
\usepackage{url}

\newcommand{\mI}{{\mathcal I}}
\newcommand{\mL}{{\mathcal L}}

\newcommand{\mS}{{\mathcal S}}

\newcommand{\mM}{{\mathcal M}}
\newcommand{\mN}{{\mathcal N}}

\newcommand{\bby}{{\bf y}}

\DeclareRobustCommand\onedot{\futurelet\@let@token\@onedot}
\def\@onedot{\ifx\@let@token.\else.\null\fi\xspace}

\def\eg{\emph{e.g}\onedot}

\definecolor{gold}{rgb}{0.85,.66,0}   

\newcommand{\revision}[1]{\textcolor{black}{#1}}

\graphicspath{{pics/}{figs/}{figs/segOverlays/}{../Miccai/figs/}{../Miccai/}{figs/figuresRev/}{figs/figures/}}

\usepackage{booktabs}
\journal{NeuroImage}

\begin{document}
\begin{frontmatter}

\title{DeepNAT: Deep Convolutional Neural Network for Segmenting Neuroanatomy}

\author{Christian Wachinger$^{a}$ \corref{cor1}} 
\cortext[cor1]{{Corresponding Author. Address: Waltherstr. 23, 81369 München, Germany; Email: christian.wachinger@med.uni-muenchen.de}}

\author{Martin Reuter$^{b,c}$, Tassilo Klein$^{d}$\corref{}}
\address{$^{a}$Department of Child and Adolescent Psychiatry, Psychosomatic and Psychotherapy, Ludwig-Maximilian-University, Munich, Germany \\
$^{b}$Athinoula A. Martinos Center for Biomedical Imaging, 
 Massachusetts General Hospital,
Harvard Medical School, Boston, MA, USA \\
$^{c}$Computer Science and Artificial Intelligence Laboratory,
Massachusetts Institute of Technology,
Cambridge, MA, USA \\
$^{d}$SAP SE, Berlin, Germany 
}

\begin{abstract}
We introduce DeepNAT, a 3D Deep convolutional neural network for the automatic segmentation of NeuroAnaTomy in T1-weighted magnetic resonance images. 
DeepNAT is an end-to-end learning-based approach to brain segmentation that jointly learns an abstract feature representation and a multi-class classification. 
We propose a 3D patch-based approach, where we do not only predict the center voxel of the patch but also neighbors, which is formulated as multi-task learning. 
To address a class imbalance problem, we arrange two networks hierarchically, where the first one separates foreground from background, and the second one identifies 25 brain structures on the foreground. 
Since patches lack spatial context, we augment them with coordinates. 
To this end, we introduce a novel intrinsic parameterization of the brain volume, formed by eigenfunctions of the Laplace-Beltrami operator. 
As network architecture, we use three convolutional layers with pooling, batch normalization, and  non-linearities, followed by fully connected layers with dropout. 
The final segmentation is inferred from the probabilistic output of the network with a 3D fully connected conditional random field, which ensures label agreement between close voxels. 
The roughly 2.7 million parameters in the network are learned with stochastic gradient descent. 
Our results show that DeepNAT compares favorably to state-of-the-art methods. 
Finally, the purely learning-based method \revision{may have} a high potential for the adaptation to young, old, or diseased brains by fine-tuning the pre-trained network with a small training sample on the target application, where the availability of larger datasets with manual annotations may boost the overall segmentation accuracy in the future. 

\end{abstract}


\end{frontmatter}

\section{Introduction}
The accurate segmentation of neuroanatomy forms the basis for volume, thickness, and shape measurements  from magnetic resonance imaging (MRI). 
Such quantitative measurements are widely studied in neuroscience to track structural brain changes associated with aging and disease. 
Additionally, they provide a vast phenotypic characterization of an individual and can serve as endophenotypes for disease.  
Since the manual segmentation of brain MRI scans is time consuming, computational tools have been developed to automatically reconstruct neuroanatomy, which is particularly important for the vastly growing number of large-scale brain studies. 
One of the most commonly used software tools for whole brain segmentation is FreeSurfer~\citep{fischl2002whole}, which applies an atlas-based segmentation strategy with deformable registration. 
This seminal work encouraged research in atlas-based segmentation, with a focus on multi-atlas techniques and label fusion strategies 
\citep{ashburner2005unified,pohl2006bmj,heckemann2006automatic,rohlfing2004evaluation,rohlfing2005quo,svarer2005mr,sabuncu2010generative,asman2012formulating,wang2013multi,wachinger2014atlas}. 
A potential drawback of atlas-based segmentation approaches is the computation of a deformation field between subjects, which involves regularization constraints to solve an ill-conditioned optimization problem.
Typically smoothness constraints are enforced, which may impede the correct spatial alignment of inter-subject scans. 
Interestingly, the deformation field is only used for propagating the segmentation and not of interest by itself. 

\revision{Learning-based approaches without deformable registration present an alternative avenue for image segmentation, where the atlas with manual segmentations serves as training set for predicting the segmentation of a new scan.}  
Directly predicting the segmentation of the entire image is challenging because of the high dimensionality, i.e., the number of voxels,  and the limited number of training scans with manual segmentations. 
Instead, the problem is reduced to predicting the label for small image regions, known as patches. 
Good segmentation performance was reported for patch-based approaches following a non-local means strategy~\citep{Coupe2011,rousseauHS11}, which is similar to a nearest neighbor search in patch space. 
Alternative patch classification schemes have been proposed, e.g., random forests~\citep{zikic2013atlas}. 
A potentially limiting factor of patch-based approaches is that they operate on image intensities, where previous results in pattern recognition suggest that it is less the classifier but rather the representation that primarily impacts the performance of a predictive model~\citep{dickinson2009object}. 
In a recent study, a wide range of image features for image segmentation was compared and a significant improvement for augmenting intensity patches with features was measured~\citep{wachinger2016descriptor}. 

While image features improve the segmentation, they are handcrafted and may therefore not be optimal for the application. 
In contrast,  neural networks autonomously learn representations that are optimal for the given task, without the need for manually defining features. 
Neural nets therefore break the common paradigm of patch-based segmentation, which separates feature extraction and classification, and replaces it with an end-to-end learning framework that starts with the image data and predicts the anatomical label. 
Deep convolutional neural networks (DCNN) have had ample success in computer vision~\citep{krizhevsky2012imagenet}  and increasingly  in medical imaging~\citep{brosch2014modeling,cirecsan2013mitosis,prasoon2013deep,roth2014new,Zheng2015,brosch2015deep,zhang2015deep,Pereira2016}. 
Applications in computer vision are typically on 2D images, where 2D+t DCNNs were proposed  for human action recognition~\citep{Ji20013}. 
In medical applications, 2.5D techniques have been proposed~\citep{prasoon2013deep,roth2014new}. 
The three orthogonal planes are integrated in existing DCNNs frameworks by setting the planes in the RGB channels. 
Difficulties in training 3D DCNNs have been reported~\citep{prasoon2013deep,roth2014new}, due to the increase in complexity by adding an additional dimension. 
Yet, several articles describe successful applications of 3D networks on medical images. 
\cite{brosch2015deep} propose a 3D deep convolutional encoder for lesion segmentation. 
\cite{Zheng2015} use a multi-layer perceptron for landmark detection. 
Most related to our work is the application of 3D convolutional neural networks, which is currently limited to few layers and  small input patches. 
\cite{li2014deep} use a 3D CNN with one convolutional and one fully connected layer for the prediction of PET from MRI on patches of $15^3$. 
\cite{brebisson2015deep} use a combination of 2D and 3D inputs for whole brain segmentation. 
The network uses one convolutional layer and 3D sub-volumes of size $13^3$. 
The foreground mask, i.e., the region that contains the labels of interest, is assumed to be given, which is not the case for scans without manual segmentation.

We propose a 3D deep convolutional network for brain segmentation that has more layers and operates on larger patches than existing 3D DCNNs, giving it the potential to model more complex relationships necessary for identifying fine-grained brain structures. 
We use latest advances in deep learning to initialize weights, to correct for internal covariate shift, and to limit overfitting for training such complex models. 
The main contributions in DeepNAT are: 
\begin{itemize}
\itemsep0em 
\item \emph{Multi-task learning:} our network does not only predict the  center label of the patch but also the labels in a small neighborhood, formulated in the DCNN as the simultaneous training of multiple tasks 
\item \emph{Hierarchical segmentation:}  we propose a hierarchical learning approach that first separates  foreground from background and then subdivides the foreground into 25 brain structures to account for the class imbalance stemming from the large background class
\item \emph{Spectral coordinates:} we introduce spectral coordinates as an intrinsic brain parameterization by computing eigenfunctions of  the Laplace-Beltrami operator on the brain mask, retaining context information in patches
\end{itemize}

The output of DeepNAT is a probabilistic label map that needs to be discretized to obtain the final segmentation. 
Performing the discretization independently for each voxel can result in spurious segmentation artifacts. 
Formulating constraints among voxels, e.g., with pairwise potentials in a random field can improve the final segmentation. 
Traditionally, such constraints have only been imposed in a small neighborhood due to computational concerns~\citep{wang2013markov}. 
We use the efficient implementation of a fully connected conditional random field (CRF) that establishes pairwise potentials on all voxel pairs~\citep{krahenbuhl2011efficient}, which was shown to substantially improve the segmentation. 
The fully connected CRF is used in combination with DCNNs for natural image segmentation in DeepLab~\citep{liang2015semantic,chen2016deeplab}. 
It is also employed for the segmentation of 2D medical images: \cite{fu2016vessel} segment vessels in 2D retinal images  and \cite{gao2016} segment the lung in  2D CT slices. 
In contrast to these approaches, we perform MAP inference of the CRF in 3D on the entire image domain to obtain the final brain segmentation.

\begin{figure*}[t]
\begin{center}
	\includegraphics[width=\textwidth]{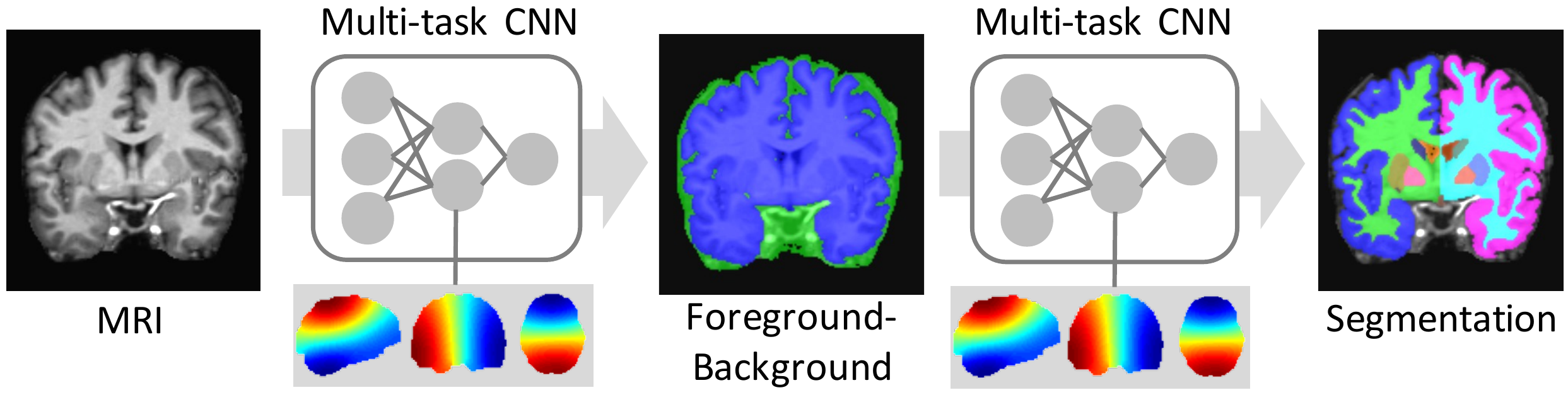} 
\caption{Overview of the hierarchical segmentation with spectral coordinates. The first multi-task DCNN separates foreground from background on skull-stripped images; the second one identifies 25 brain structures on the foreground. 
\label{fig:overview}
}
\end{center}
\end{figure*}

\begin{table*}
 \centering
  \begin{tabular}{llll}
  \toprule
\multicolumn{4}{c}{3D Multi-Task Network Architecture} \\  
Layers & Specification & Layers & Specification \\
  \midrule
1. Convolution	&	$7 \times 7 \times 7 \times 32$  & 10. Inner Product	&	Neurons: 1024\\
2. ReLU	&		& 11. ReLU	&\\
3. Max-Pooling	&	Size: 2, Stride: 2 \quad \quad \quad & 12. Dropout	&	Rate: 0.5	\\
4. Convolution	&	$5 \times 5 \times 5 \times 64$ & 13. Concatenation	&	w/ coordinates 	\\
5. Batch normalization &  & 14. Inner Product	&	Neurons: 512\\
6. ReLU	&		& 15. Batch normalization &\\
7. Convolution	&	$3 \times 3 \times 3 \times 64$ & 16. ReLU	&	\\
8. Batch normalization & & 17. Dropout	&	Rate: 0.5 \\
9. ReLU	&		 Output: $1728$ & 18. Inner Product ($\times$ tasks)	&	Neurons: 2 / 25	 \\
    \bottomrule
 \end{tabular}
\caption{DeepNAT network architecture with convolutional (left) and fully connected (right) parts. The size of the input patch is $23^3$. The last layer is replicated for multi-task learning according to the number of tasks. The cascaded networks are identical except for the number of neurons in the last layer.  }
 \label{tab:archi} 
\end{table*}

\begin{table*}
 \centering
  \begin{tabular}{lcrcc}
  \toprule
  Layer & Parameter calculation & \# Parameters & Input Dimensionality & Output Dimensionality \\
  \midrule
Convolution I	&	$7 \times 7 \times 7 \times 1 \times 32$	&	10,976	&	$23 \times 23 \times 23 \times 1$	&	$17 \times 17 \times 17 \times 32$	\\
Convolution II	&	$5 \times 5 \times 5 \times 32 \times 64$	&	256,000	&	$9 \times 9 \times 9 \times 32$	&	$5 \times 5 \times 5 \times 64$	\\
Convolution III	&	$3 \times 3 \times 3 \times 64 \times 64$	&	110,592	&	$5 \times 5 \times 5 \times 64$	&	$3 \times 3 \times 3 \times 64$	\\
Inner Product I	&	$3 \times 3 \times 3 \times 64 \times 1024$	&	1,769,472	&	$3 \times 3 \times 3 \times 64$	&	1024	\\
Inner Product II	&	$(1024+6) \times 512$	&	527,360	&	1030	&	512	\\
Inner Product III	&	$512 \times 25$	&	12,800	&	512	&	25	\\
    \bottomrule
 \end{tabular}
\caption{The number of parameters to be learned at each of the convolutional and inner product layers of the network. The total number of parameters is 2,687,200. We also state the input and output dimensionality of each of the layers, which provides insights about the internal representation. Note that max-pooling operates before convolution II and that spatial coordinates are concatenated before inner product II. The list does not include bias parameters, which are negligible in size.}
 \label{tab:params} 
\end{table*}

\section{Method}
Given a novel image $I$, we aim to infer its segmentation~$S$ based on training images $\mI = \{ I^1, \ldots, I^n\}$ with segmentations $\mS = \{S^1, \ldots, S^n\}$. 
A probabilistic label map $\mL = \{L^1, \ldots, L^\eta\}$ specifies the likelihood for each brain label~$l \in \{1, \ldots, \eta\}$%
\begin{align}
L^l(x) \ = \ p( S(x) = l | I; \mI, \mS). %
\end{align}
Let $I(\mN_x)$ denote an image patch centered at location $x$, the likelihood in a patch-based segmentation approach is
\begin{align}
L^l(x) \ = \ p( S(x) = l | I(\mN_x); \mI, \mS). \label{eq:prob}
\end{align}
We estimate the likelihood by training a deep convolutional neural network, where the patch inference corresponds to multi-class classification. 
\revision{We skull strip the images to focus the prediction on the brain mask; a brain scan from which the skull and other non-brain tissue like dura and eyes are removed.}

\subsection{Hierarchical Segmentation}
Figure~\ref{fig:overview} illustrates the hierarchical approach for whole brain segmentation in DeepNAT. %
\revision{In the first cascade, brain regions are classified into foreground and background. 
The foreground consists of 25 major brain structures that are illustrated in Figure~\ref{fig:segResults}. 
The background is the region within the brain mask that is not part of the foreground.} 
Data that is classified as foreground undergoes the next cascaded step to identify separate brain structures. 
Given the inherent class imbalance, the hierarchical segmentation has the potential to perform better than a single-step classification, which classifies into brain regions as well as background. %
Problems with a large background class have previously been noted for atlas-based segmentation~\citep{wachinger2014atlas}.
The background is typically represented by a large pool of data, while small brain structures are prone to being underrepresented. %
On our data, we measured a foreground to background volume ratio of about 2 to 1. 
The background is therefore substantially larger than any of the individual brain structures on the foreground.  
As data augmentation allows only for crude and poor compensation, the cascaded approach presents a viable alternative. %

\subsection{Network Architecture \label{sec:arch}}

Multi-layer convolutional neural networks pioneered by~\cite{LeCun:1989:BAH:1351079.1351090} have led to breakthrough results, constituting the state-of-the art technology for many challenges such as ImageNet~\citep{krizhevsky2012imagenet}. The underlying idea is to create a deep hierarchical feature representation  that shares filter weights across the input domain.  This allows for the robust modeling of complex relationships while requiring a reduced number of parameters, for which solutions can be obtained by stochastic gradient descent.

Table~\ref{tab:archi} lists the details of the DeepNAT network architecture, where both networks (for each cascade) are identical except for the number of neurons in the last layer (2 and 25, respectively). 
The network consists of three convolutional layers, where in each layer the filter masks are to be learned. 
A filter mask is specified by the spatial dimension, \eg, $5 \times 5 \times 5$ and the number of filters to be used, \eg, 64. 
Each filter extends to all of the input channels. As an example, the filters are of size $5 \times 5 \times 5 \times 32$ in the second convolution. 
The total number of free parameters to be estimated is the filter size times the number of filters, so $5 \times 5 \times 5 \times 32 \times 64$ for the second convolution. 
Table~\ref{tab:params} states the number of parameters together with the input and output dimensionality for each layer. 
Note that for 2D DCNNs the filters have 3 dimensions, whereas for 3D DCNNs the filters have 4 dimensions. %

Each convolution is followed by a rectified linear unit (ReLU)~\citep{hahnloser2003permitted,glorot2011deep}, which supports the efficient training of the network with reduced risk for vanishing gradient compared to other non-linearities. 
The aim of the convolutional part of the network is to reduce the dimensionality from the initial patch size of $23 \times 23 \times 23$ before entering the fully connected stage. 
Although each convolution reduces the size, we use an additional max-pooling layer with stride two to arrive at a $3^3$ block of neurons at the end of the convolutional stage.
The $3^3$ block is an explicit design choice. 
A smaller $2^3$ block would cause a lack of localization, with the patch center being split into exterior blocks. 
A larger $4^3$ block would dramatically increase the number of parameters at the end of the convolutional stage, where most free parameters occur at the intersection between convolutional and fully connected layers, see Table~\ref{tab:params}.  

We use batch normalization at several layers in the network to reduce the internal covariate shift~\citep{ioffe2015batch}. 
It accounts for the problem that the distribution of each layer's inputs changes during training, as the parameters of the previous layers change, which is more pronounced in 3D networks. 
We further use two dropout layers, which randomly disable neurons in the network. 
This helps with the generalizability of the network by acting as a regularizer and mitigating overfitting.
To resolve potential location ambiguity, coordinates of the patches are given to the network, see Sec.~\ref{sec:spectCoord}. 
This is achieved by concatenating the image content after the first fully connected layer with the location information in layer 13. 
In the training stage, we compute the multinomial logistic loss as last layer, where the probability distribution over classes is inferred from the last inner product layer with a softmax.  

For the initialization of the weights, we use the Xavier algorithm that automatically determines the scale of initialization based on the number of input and output neurons~\citep{glorot2010understanding}. This initialization supports training deep networks without requiring per-layer pre-training because signals can reach deep into the network without shrinking or growing too much.

\subsection{Multi-task Learning} 

In Eq.(\ref{eq:prob}), we use an image patch to predict the tissue label of the center voxel. Performing this inference on the entire image results in a single vote per voxel. 
Previous results in patch-based segmentation have, however, demonstrated the advantage of propagating not only the center label but also neighboring labels~\citep{rousseauHS11}. 
With such an approach, the voxel label is not only inferred from a single patch, but also from neighboring patches. 
\cite{rousseauHS11} refer to this as the multi-point  method in the context of non-local means segmentation. 
We propose to replicate the multi-point method for DCNN segmentation by employing multi-task learning. 
Instead of learning a single task, which predicts the center label, we simultaneously learn multiple tasks, which predict the center and surrounding neighborhood. 
The neighborhood size determines the number of tasks. 
While there have been applications for deep multi-task learning~\citep{yim2015rotating,luong2015multi}, we are not aware of previous applications for image segmentation. 

We implement multi-task learning in the CDNN architecture by replicating the last inner product layer (\#18) according to the number of tasks. 
The increase in the number of parameters to be learned is limited by this setup because all tasks share the same network, except for last inner product layer that specializes on the task.  
Each task~$t$ predicts the likelihood $p_t( S(x_t) = l | I(\mN_x); \mI, \mS)$ for locations $x_t$ in the neighborhood $\mM_x$ centered around $x$. 
We compute the multi-task likelihood for the label by averaging likelihoods across tasks 
\begin{align}
L^l(x) \ = \ \frac{1}{| \mM_x |} \sum_{x_t \in \mM_x} p( S(x) = l | I(\mN_{x_t}); \mI, \mS). \label{eq:mult}
\end{align}
We experiment with 7 and 27 neighborhood systems $\mM$  for the prediction, where the 7 neighborhood consists of the 6 direct neighbors and the 27 neighborhood consists of the full $3^3$ region. 
\revision{From a different perspective, this approach of averaging among multiple predictions per voxel can also be seen as an ensemble method.}

\subsection{Spectral Brain Coordinates \label{sec:spectCoord}}

A downside of patch-based segmentation techniques is the loss of spatial context~\citep{wachinger2016descriptor}. 
Considering the symmetry of the brain, it is easy to confuse patches across hemispheres. 
In addition, context provides valuable information for structures with low tissue contrast. 
To increase the spatial information, we augment patches with location information. 
Previous approaches have, for instance, used Cartesian coordinates~\citep{wachinger2014importance} or distances to centroids~\citep{brebisson2015deep}. %
We propose spectral brain coordinates as an alternative parameterization of the brain volume, which we obtain by computing eigenfunctions of the Laplace-Beltrami operator inside the 3D brain mask. 
Eigenfunctions of the cortex surface have previously been used for brain matching~\citep{lombaert2013diffeomorphic,lombaert2013focusr} and eigenvalues as shape descriptors~\citep{Wachinger2015brainprint}.
In contrast, we compute spectral coordinates on the solid (volume) and use it as an intrinsic coordinate system for learning.
On the brain mask, we solve the Laplacian eigenvalue problem 
\begin{equation}
\Delta f = - \lambda f 
\end{equation} 
with the Laplace-Beltrami operator~$\Delta$, eigenvalues~$\lambda$ and eigenfunctions~$f$. 
We approximate the Laplace-Beltrami operator with the graph Laplacian~\citep{chung1997spectral}. 
The weights in the adjacency matrix $W$ between two points $i$ and $j$ are set to 1 if both points are neighbors and within the brain mask, otherwise they are set to 0. 
This yields a sparse matrix $W$. 
The Laplacian operator on a graph is 
\begin{align}
L = D  - W \quad \quad \quad \quad D_{ii} = \sum_j W_{ij}
\end{align}
with the node degree matrix $D$. %

\begin{figure}[t]
\begin{center}
	\includegraphics[width=0.5\textwidth]{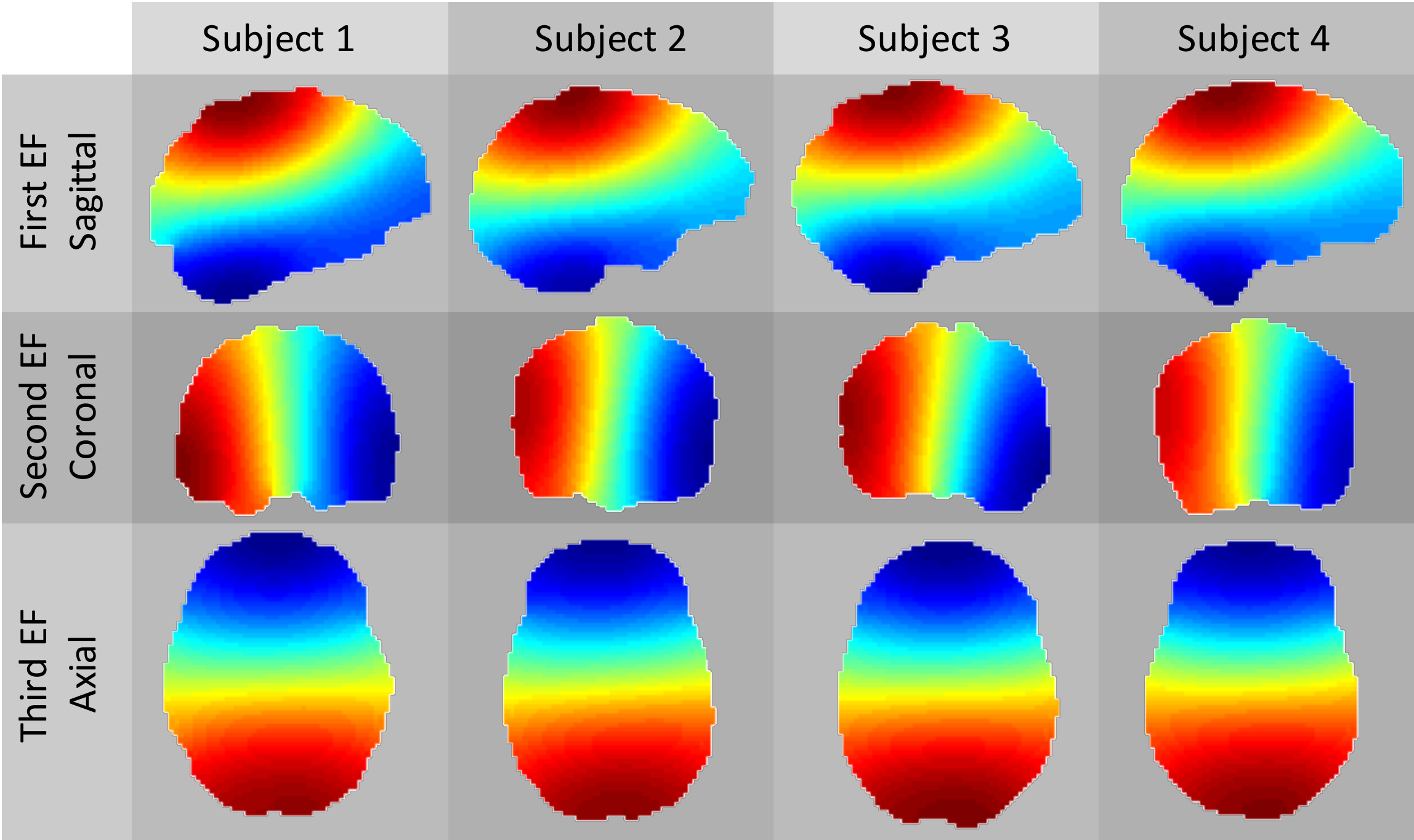} 
\caption{Illustration of first three eigenfunctions (EF) for four subjects. Each function is shown on the anatomical view that best highlights the gradient. %
\label{fig:spect}
}
\end{center}
\end{figure}

We compute the first three non-constant eigenvectors of the Laplacian, where each eigenvector corresponds to a 3D image and the ensemble of eigenvectors forms the spectral brain coordinates. 
Fig.~\ref{fig:spect} illustrates the first three eigenvectors, which roughly represent vibrations along primary coordinate axes. 
The consistency of the coloring across the four subjects highlights the potential for an accordant encoding of location information. 
Note that the eigenvectors are isometry invariant to the object, meaning that they do not change with rotations or translations. 
Hence they present an intrinsic parameterization independent of the brain orientation or location. 
This independence can be seen from the graph construction encoded in the adjacency matrix. \revision{The adjacency structure only depends on neighborhood relationships, which do not change with image  translation or rotation.}

Depending on the object to parameterize and the number of eigenfunctions, flipping due to sign ambiguity or swapping of eigenfunctions may hinder a direct comparison. 
\cite{lombaert2013focusr} proposed an approach for spectrum ordering. 
In our application, with only computing the first three eigenfunctions of the brain mask, no correction was required. 
Note that we could also compute more than three eigenfunctions to increase the amount of spatial information in the DCNN, which may require a re-ordering strategy. 
To the best of our knowledge, this is the first application of eigenfunctions of the 3D solid for defining an intrinsic brain coordinate system. 

Following the idea of providing the neural net with all the data and letting it pick the relevant information, we input next to the three spectral coordinates also the three Cartesian coordinates. We normalized the Cartesian coordinates, by subtracting the center of mass of the brain mask to make them more comparable across scans.

\subsection{Fully Connected Conditional Random Field}
The DCNN prediction results in a probabilistic brain segmentation. 
To obtain the final segmentation, we use maximum a posteriori inference on a conditional random field (CRF). 
The CRF allows for formulating potentials that ensure label agreement between close voxels with smoothness terms and follow the image content with appearance terms. 
Traditionally, short-range CRFs with connections between neighboring locations have been used~\citep{wang2013markov}, which can however yield excessive smoothing of organ boundaries. 
In contrast, the fully connected CRF defines pairwise potentials on all pairs of image locations. 
The vast number of pairwise potentials to be defined makes conventional inference impractical. 
We use the highly efficient approximate inference algorithm proposed by \cite{krahenbuhl2011efficient}  to infer a fully connected CRF model on the entire 3D brain. 
Key for the efficient computation is the definition of pairwise edge potentials by a linear combination of Gaussian kernels. 

\begin{figure*}[t]
\begin{center}
	\includegraphics[width=0.4\textwidth]{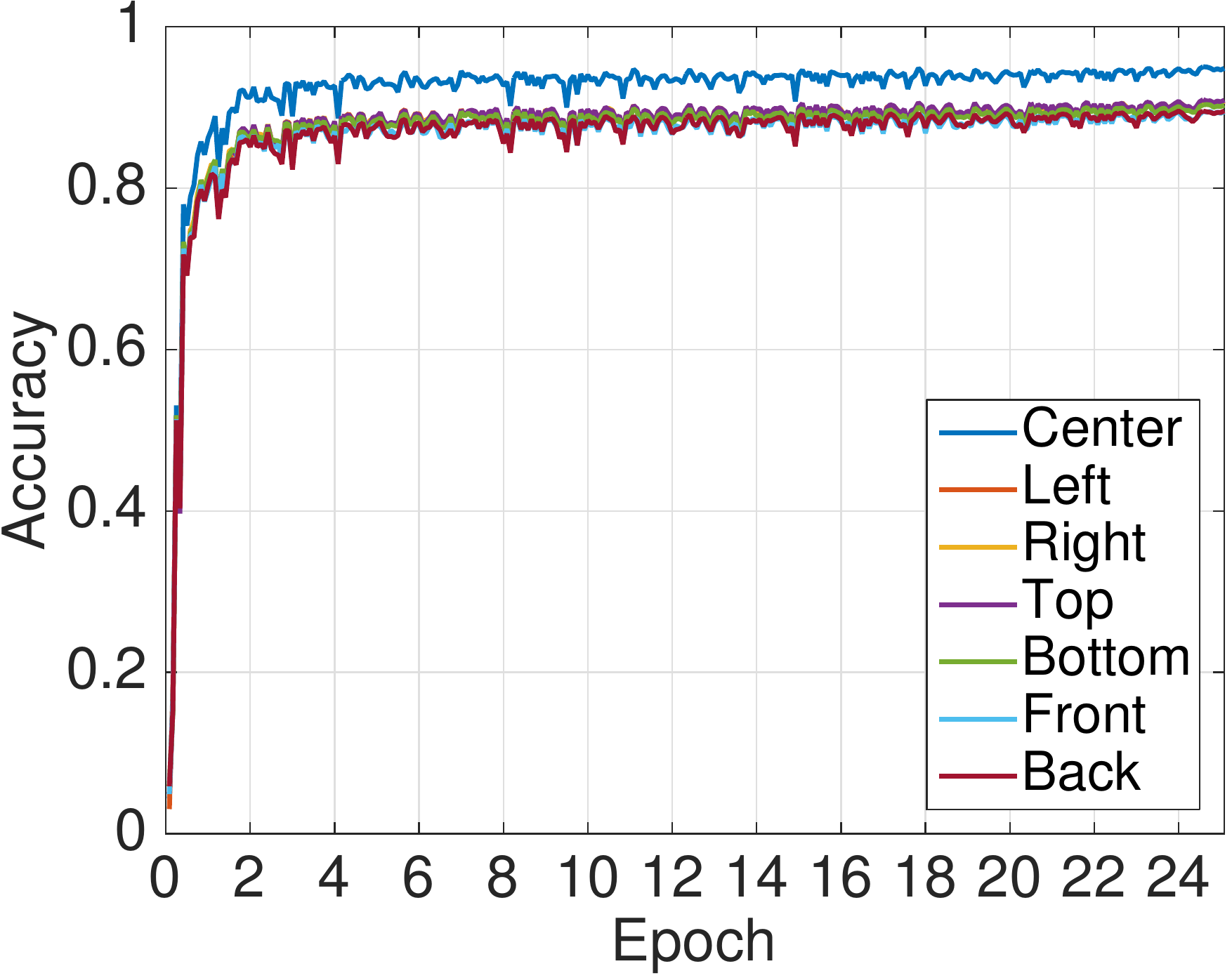}  \quad \quad
	\includegraphics[width=0.39\textwidth]{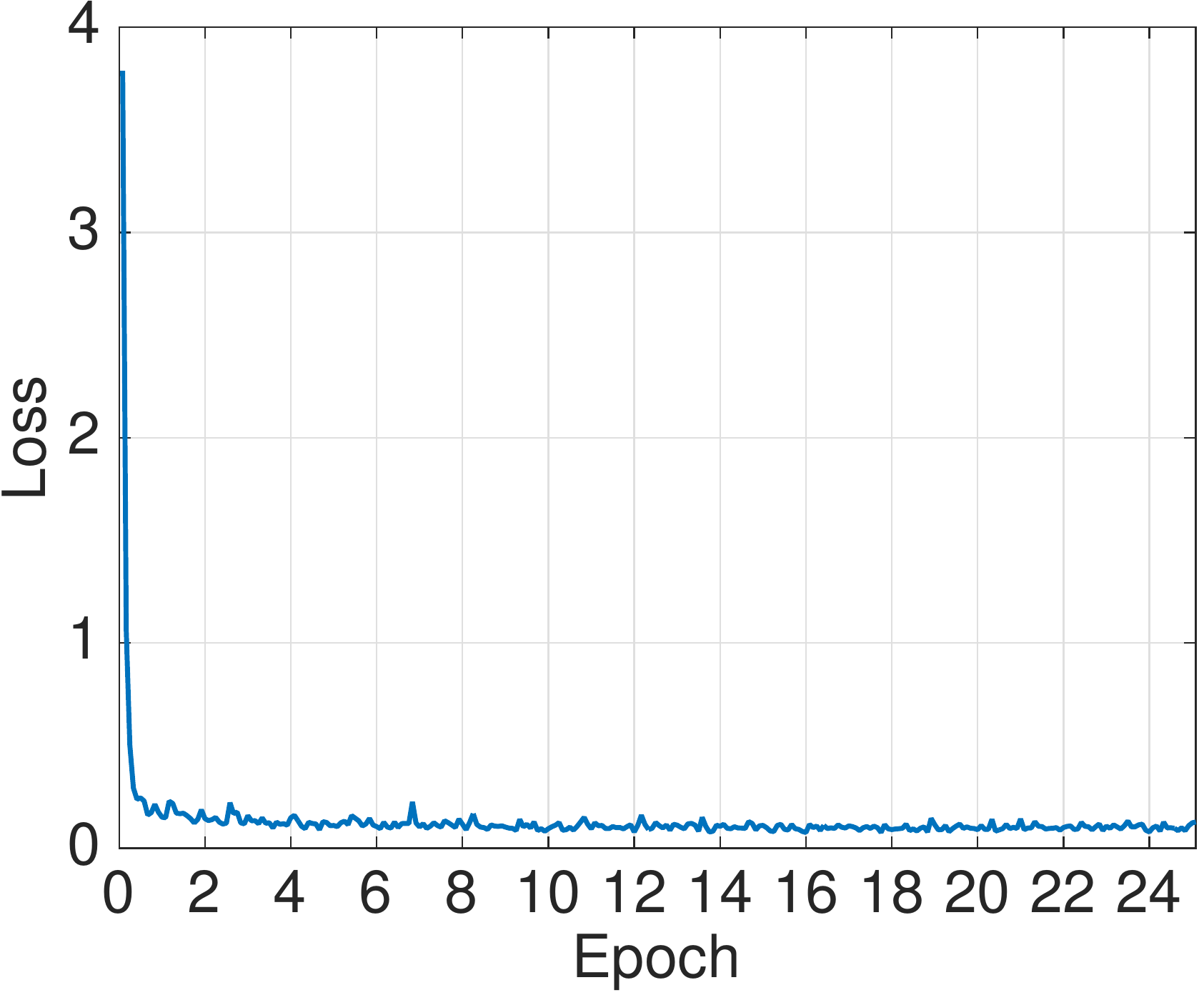} 
\caption{Accuracy and loss for DeepNAT during training for 25 epochs. The accuracy is shown for all seven tasks, which predict the label of the center voxel and neighbors.  
\label{fig:epochsLoss}}
\end{center}
\end{figure*}

The inference algorithm uses mean field approximation that is iteratively optimized with a series of message passing steps. 
Importantly, the message passing updates for a fully decomposable mean field approximation is identical to Gaussian filtering in bilateral space. 
With the help of efficient approximate high-dimensional filtering~\citep{adams2010fast}, the computational complexity of message passing is reduced from quadratic to linear in the number of variables.

The Gibbs energy of the CRF model is
\begin{equation}
E(\bby) = \sum_i \psi_u(y_i) + \sum_{i \leq j} \psi_p(y_i,y_j),
\end{equation}
with the label assignment~$\bby$ and $i$, $j$ ranging from 1 to the number of voxels. 
The unary potential $\psi_u(y_i) = - \log P(y_i)$ is defined as the negative log likelihood of the label assignment probability from the multi-task DCNN in Eq.(\ref{eq:mult}). 
We use the pairwise potential from~\citep{krahenbuhl2011efficient}, which allows for efficient inference on fully connected graphs. 
Given image intensities $I_i$ and $I_j$ with locations $p_i$ and $p_j$, the pairwise potential is
\begin{align}
\psi_p (y_i,y_j) &= \mu_{ij} \left[ v_1 e^{ -\frac{ \| p_i - p_j \|^2}{2 \sigma_\alpha^2}  -\frac{ ( I_i - I_j )^2}{2 \sigma_\beta^2}  } +  v_2 e^{-\frac{ \| p_i - p_j \|^2}{2 \sigma_\gamma^2}  } \right].
\end{align}
The first exponential term models the appearance where nearby voxels with similar intensity are likely to show the same structure, controlled by  spatial $\sigma_\alpha$ and intensity $\sigma_\beta$ parameters; this corresponds to a bilateral kernel. 
The second exponential term models the smoothness by considering spatial proximity, controlled by~$\sigma_\gamma$. 
The appearance and smoothness terms are weighted by parameters, $v_1$ and $v_2$, respectively. 
For the label compatibility the Potts model is used, $\mu_{ij} = [ y_i \neq y_j ]$.

\section{Results}

We evaluate the segmentation on the dataset of the MICCAI Multi-Atlas Labeling challenge\footnote{https://masi.vuse.vanderbilt.edu/workshop2012}~\citep{landman2012miccai}, which consists of T1-weighted MRI scans from 30 subjects of OASIS~\citep{marcus2007open}. 
Manual segmentations were provided by Neuromorphometrics, Inc.\footnote{http://Neuromorphometrics.com/} under academic subscription. 
The images are 1 mm isotropic with a slice size of $256 \times 256$ pixels and the number of slices varying above 256. 
To improve the estimation of the roughly 2.7 million parameters in the network, we increase the number of training scans from 15 in the challenge to 20. 
The remaining 10 scans are used for testing. 
We compare our results to PICSL~\citep{wang2012challenge}, the winner of the MICCAI labeling challenge that uses deformable registration, label fusion, and corrective learning. 
In addition, we compare to spatial STAPLE~\citep{asman2012formulating}, which is an extension of the popular simultaneous truth and performance level estimation (STAPLE) method~\citep{warfield2004simultaneous}. 
It is among the best performing methods in the challenge and allows for a spatially varying performance of raters, i.e., registered atlas. 
Finally, we compare to the segmentation with FreeSurfer v5.3~\citep{dale1993improved,dale1999cortical,fischl1999cortical,fischl1999high,fischl2002whole}. 
In contrast to the other methods, FreeSurfer comes with its own atlas and does not use the training data. 
\revision{We measure the segmentation accuracy with the Dice volume overlap score~\citep{dice1945measures} between the automatic segmentation~$S$ and manual segmentations~$\bar{S}$ 
\begin{equation}
D(S, \bar{S}) = 2 \frac{ | S \cap \bar{S} | } {|S| + |\bar{S}|}.
\end{equation}}

\begin{figure}[ht!]
\begin{center}
	\includegraphics[width=0.4\textwidth]{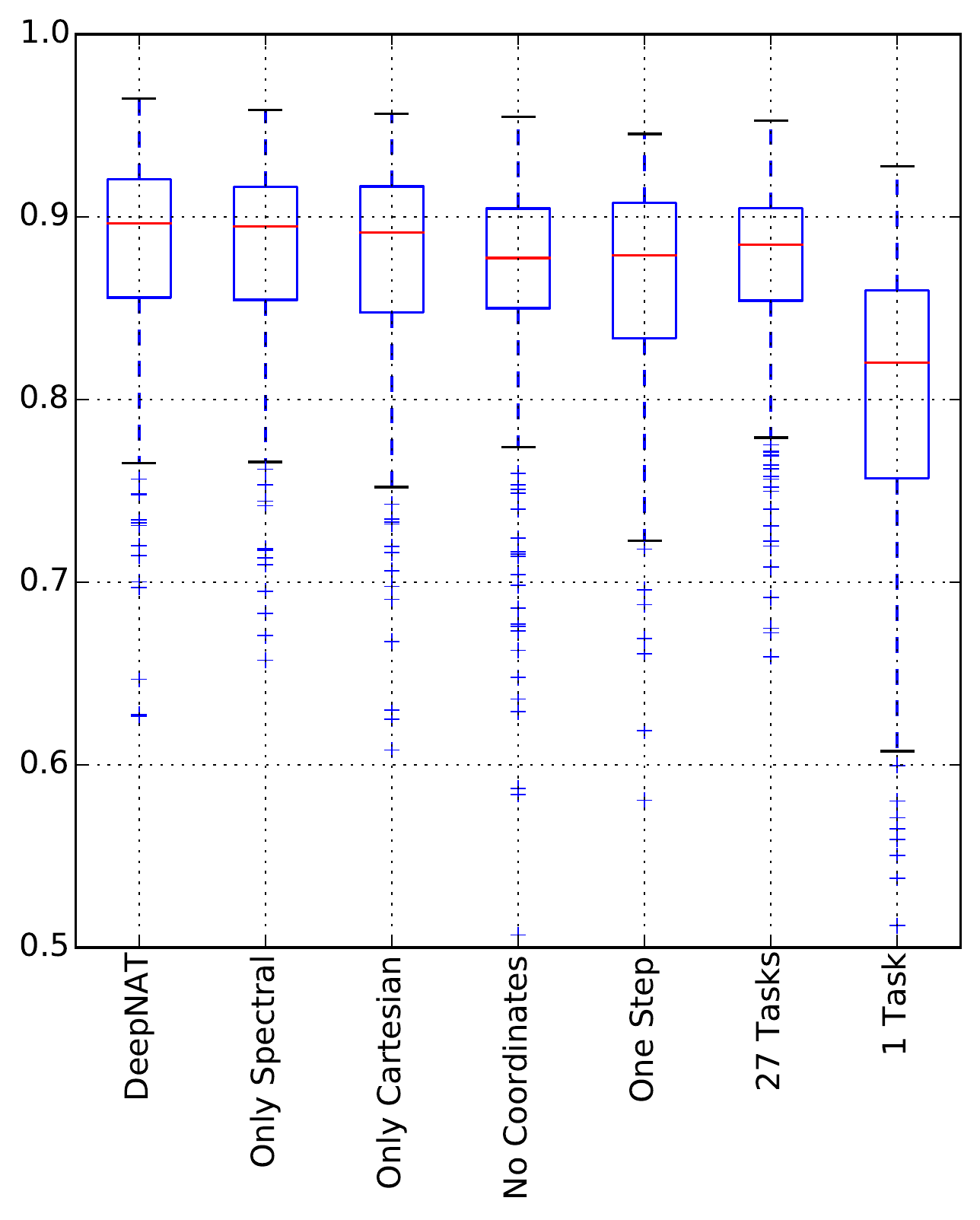} 
    \begin{tabular}{lr}
    Method & Median Dice \\ 
    \toprule
    DeepNAT &  \textbf{0.897} \\ 
    \midrule
    Only Spectral  &  0.895  \\
    Only Cartesian & 0.892 \\ 
    No Coordinates & 0.878 \\ 
    \midrule
    One Step & 0.879 \\ 
    \midrule
    27 Tasks & 0.885 \\ 
    1 Task & 0.820 \\
  \end{tabular}	  
\caption{Segmentation results in Dice for different configurations of DeepNAT. In the figure, red line indicates the median, the boxes extend to the $25^{\text{\tiny th}}$ and $75^{\text{\tiny th}}$ percentiles, and the whiskers reach to the most extreme values not considered outliers (crosses). 
The table lists the median Dice for the different variations in the DeepNAT configuration. 
\label{fig:boxMethod}
}
\end{center}
\end{figure}

\begin{figure}[t]
\begin{center}
	\includegraphics[width=0.4\textwidth]{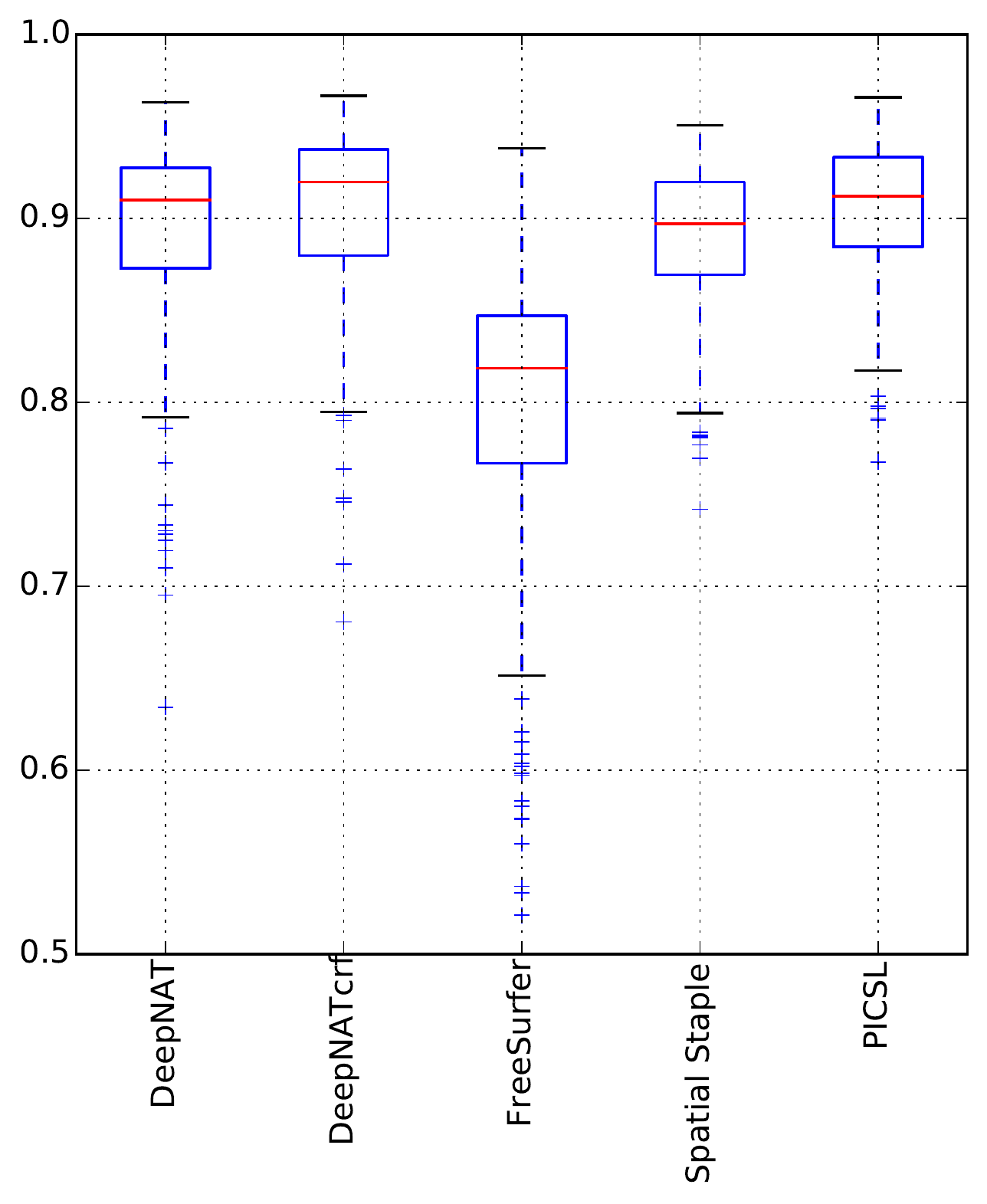} 
    \begin{tabular}{lr}
    Method & Median Dice \\ 
    \toprule
    DeepNAT &  0.910 \\
    DeepNATcrf &  \textbf{0.920}  \\
    FreeSurfer & 0.819 \\ 
    Spatial STAPLE & 0.897 \\ 
    PICSL & 0.912 \\
  \end{tabular}	  
\caption{Segmentation results in Dice for DeepNAT and DeepNATcrf  together with alternative segmentation methods: FreeSurfer, spatial STAPLE, and PICSL. In the figure, red lines indicate the median, the boxes extend to the $25^{\text{\tiny th}}$ and $75^{\text{\tiny th}}$ percentiles, and the whiskers reach to the most extreme values not considered outliers (crosses).  
The table lists the median Dice for the different approaches. 
Note that the results for DeepNAT vary from Fig.~\ref{fig:boxMethod} due to longer training. 
\label{fig:boxComp}
}
\end{center}
\end{figure}

\begin{figure}[t]
\begin{center}
	\includegraphics[width=0.4\textwidth]{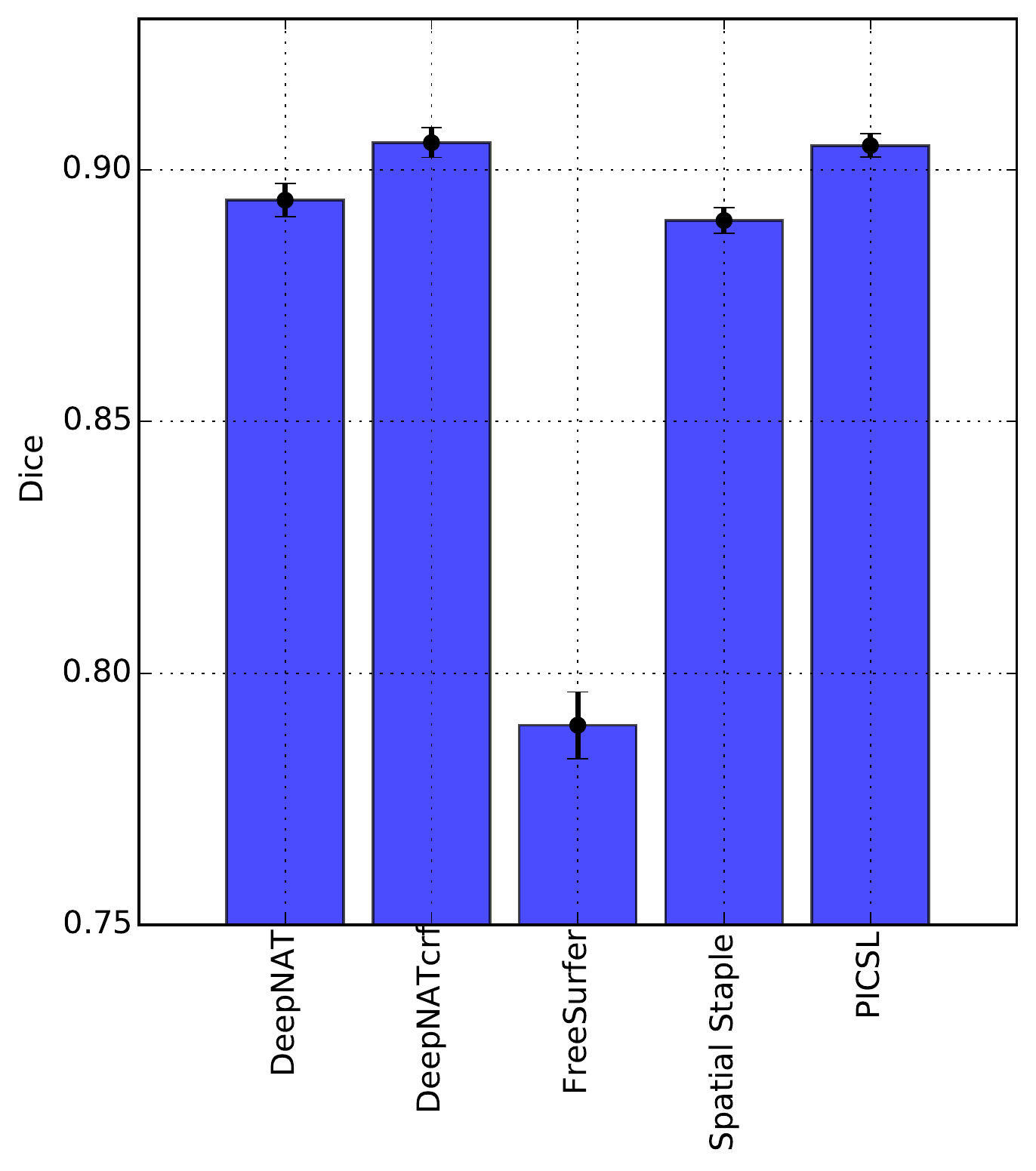} 
    \begin{tabular}{lr}
    Method & Mean Dice \\ 
    \toprule
    DeepNAT &  0.894 \\
    DeepNATcrf &  \textbf{0.906}  \\
    FreeSurfer & 0.790 \\ 
    Spatial STAPLE & 0.890 \\ 
    PICSL & 0.904 \\
  \end{tabular}	  
\caption{\revision{Segmentation results in Dice for DeepNAT and DeepNATcrf  together with alternative segmentation methods: FreeSurfer, spatial STAPLE, and PICSL. 
The bars show the mean Dice score and the lines show the standard error. 
The table lists the mean Dice for the different approaches. } 
\label{fig:boxCompMean}
}
\end{center}
\end{figure}

We select a patch size of $23 \times 23 \times 23$ as a trade-off between a large enough image region for the label classification and memory consumption as well as processing speed. 
DeepNAT is based on the Caffe framework~\citep{jia2014caffe}.
Gradients are computed on minibatches, where each gradient update is the average of the individual gradients of the patches in the minibatch. 
The size of the minibatch is constrained by the memory of the GPU, where a size of 2,048 fills up most of the 12GB GPU memory on the NVIDIA Tesla K40 and TITAN X used in the experiments. %
Large batch sizes are advisable as they better approximate the true gradient.

\begin{figure*}[t]
\begin{center}
	\includegraphics[width=\textwidth]{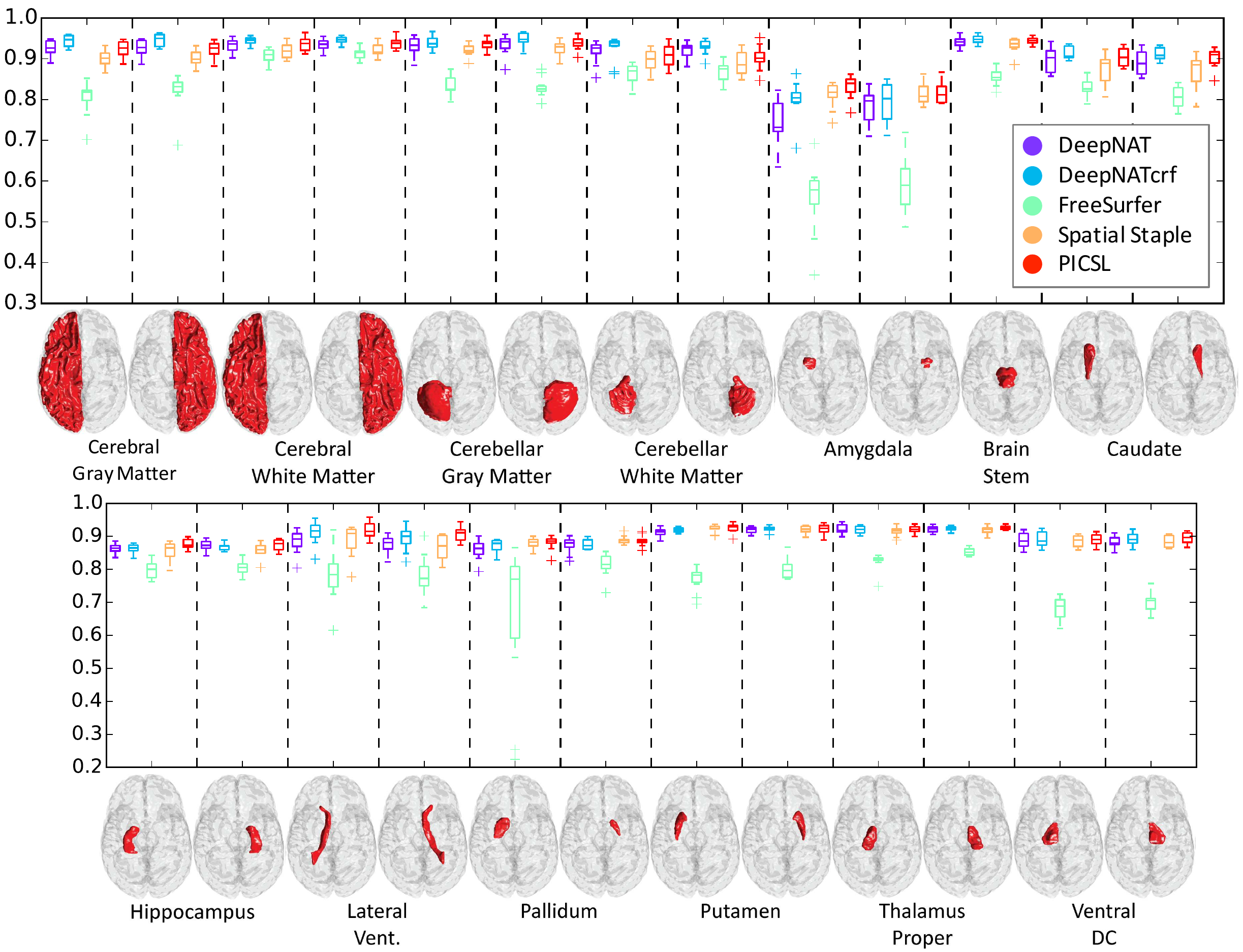}
\caption{Segmentation results in Dice for DeepNAT, DeepNATcrf, FreeSurfer, spatial STAPLE, and PICSL for 25 brain structures.  Centerline indicates the median, the boxes extend to the $25^{\text{\tiny th}}$ and $75^{\text{\tiny th}}$ percentiles, and the whiskers reach to the most extreme values not considered outliers (crosses).
We are grateful to Bennett Landmann for 3D renderings of brain structures. 
\label{fig:segResults}}
\end{center}
\end{figure*}

We train the network with stochastic gradient descent and the ``poly" scheme (also applied by~\cite{chen2016deeplab}) using a base learning rate of 0.01. %
The actual learning rate at each iteration is the base learning rate multiplied by $(1 - \text{iteration} / \text{max\_iteration} )^{0.9}$, promoting larger steps at the beginning of the training period and smaller steps towards the end. 
For the first network, we randomly sample 30,000 patches from the foreground and background in each training image, yielding 1.2 million training patches. 
For the second network, we randomly sample at most 3,000 patches per structure, where we double the number of patches for the white matter and gray matter to account for the higher variability in these classes, yielding a total of about 1.1~million training patches.   
We apply inhomogeneity correction and intensity normalization from the FreeSurfer pipeline to the MRI scans. 
In light of a small number of training images with manual segmentations, the standardization yields higher homogeneity in the dataset and should therefore facilitate the inference task. 
We set the CRF parameter to standard settings~$v_1 = v_2 = 3$, $\sigma_\alpha = \sigma_\gamma = 3$, and $\sigma_\beta = 10$~\citep{chen2016deeplab}. 
\revision{Figure~\ref{fig:epochsLoss} shows the accuracy and loss during training for the second network.} 
For the accuracy, we have a different line for each of the seven tasks. 
Notably, the center task achieves the highest accuracy, where the remaining tasks which predict labels for neighboring voxels show comparable results. 
This is insofar surprising that all tasks have the same weight in the network and it suggests that it is intrinsically easier for the network to predict the patch center. 
Overall, we observe a fast convergence to a relatively high classification results, where prolonged training yields a small but steady improvement of the accuracy. 

\begin{figure*}[t]
\begin{center}
	\includegraphics[width=\textwidth]{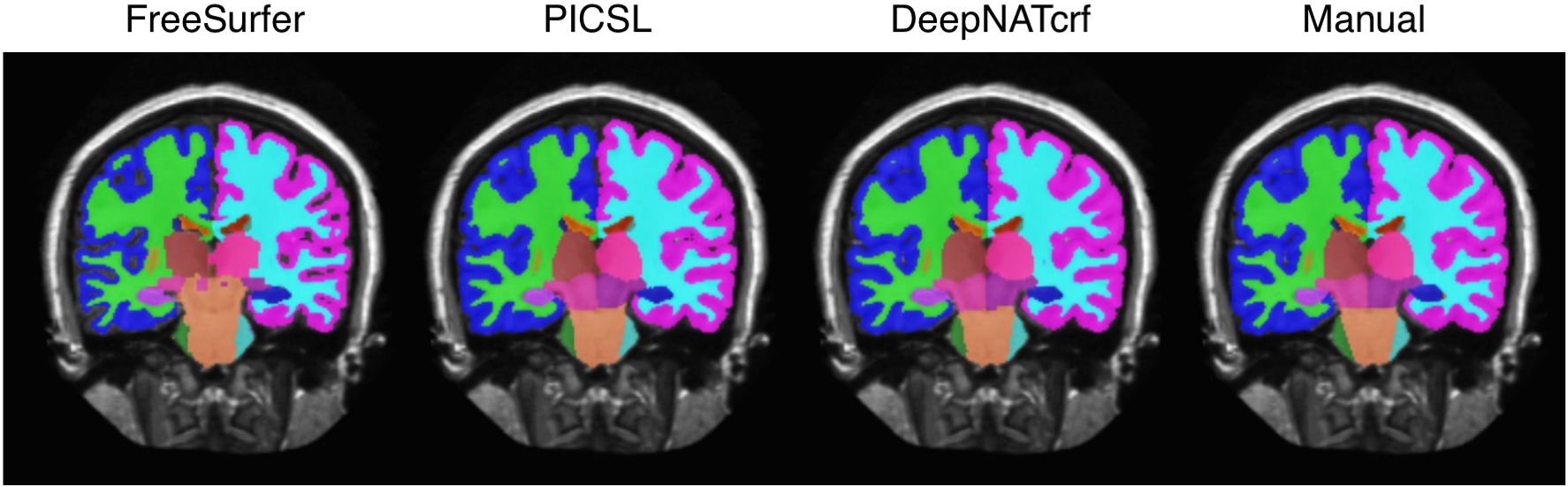} 
\caption{Example segmentations for FreeSurfer, PICSL, and DeepNATcrf together with the corresponding manual segmentation. \revision{FreeSurfer shows the largest variations with respect to the manual segmentation, particularly in cortical structures and the brainstem. The results of PICSL and DeepNATcrf are highly similar to the manual segmentation.} 
\label{fig:slices}
}
\end{center}
\end{figure*}

First, we evaluate the impact of the proposed contributions in DeepNAT on the segmentation accuracy: (i)~coordinates, (ii)~hierarchical architecture, and (iii)~multi-task learning. 
We perform the comparison by using the DeepNAT network, which uses seven tasks and combines spectral and Cartesian coordinates. 
We modify one of the network settings while  keeping the remaining configuration. 
Figure~\ref{fig:boxMethod} shows the segmentation results, where the statistics are computed across all of the 25  brain structures. 
Each setting is trained for 8 epochs, which takes about 1 day. 
The segmentation of a new scan at test time takes about 1 hour. 
With respect to coordinates, we observe a clear drop when using no coordinates. 
Spectral coordinates perform slightly better than Cartesian coordinates, where the combination of both in DeepNAT yields the highest accuracy. 
Next we compare the hierarchical approach to directly segmenting the 25 structures in one step, where the one step approach yields a lower accuracy. 
Finally, we evaluate the importance of multi-task learning. 
We compare with single task prediction, which only predicts the center voxel of the patch, and with the prediction of a larger number of tasks, 27. 
\revision{The results show in comparison to the seven tasks in DeepNAT  a strong decrease in accuracy for the single task and a small decrease in accuracy for 27 tasks.  
We test the significance of DeepNAT to each of the variants with the pairwise non-parametric Wilcoxon signed-rank test (two-sided). 
The improvement of DeepNAT over only spectral is significant with $p<0.05$ and the improvement over all other variants is significant with $p<0.001$.} 

\revision{We further evaluated different parameters for the optimization of the network. The reduction of the base learning rate to 0.005 leads to a median Dice score of 0.888. 
The usage of a minibatch size of 512 yields a median Dice score of 0.895. 
The application of the ADAGRAD~\citep{duchi2011adaptive} stochastic optimization results in a median Dice score of 0.881, compared to 0.897 in DeepNAT.}

For the second evaluation, we train DeepNAT for 25 epchos, which took about 3 days and compare it to alternative segmentation approaches: FreeSurfer, spatial STAPLE, and PICSL. 
\revision{Figures~\ref{fig:boxComp} shows the results over all 25 brain structures with the median and percentiles and Figure~\ref{fig:boxCompMean} shows the mean and standard error.}
DeepNATcrf denotes the estimation of the final segmentation with the fully connected CRF, where for DeepNAT we infer the segmentation independently for each voxel with  weighted majority voting. 
 The mean and median Dice for DeepNAT is higher than for FreeSurfer or spatial STAPLE. 
The CRF yields an increase in Dice by about 0.01 and the overall highest segmentation accuracy. 
Figure~\ref{fig:segResults} shows detailed results for all of the 25 brain structures.

\revision{Across all structures, DeepNATcrf yields significantly higher Dice scores in comparison to DeepNAT ($p < 0.001$), FreeSurfer ($p < 0.001$), and STAPLE ($p < 0.001$). 
The difference to PICSL ($p=0.27$) is not significant.}  
We further explore the difference between PICSL and DeepNATcrf on a per structure basis. 
\revision{Here DeepNATcrf yields significantly higher values for left cerebral gray matter ($p < 0.005$), right cerebral gray matter ($p < 0.005$),  right cerebral white matter ($p < 0.05$), right cerebellar white matter ($p < 0.05$), and left caudate ($p < 0.05$)
while PICSL yields significantly higher values for left amygdala ($p < 0.01$), right caudate ($p < 0.05$), and left hippocampus ($p < 0.01$). 
The different results for the left and right caudate are due to variations in median Dice in PICSL (left: 0.903, right 0.910) compared to more consistent results across hemispheres for DeepNATcrf (left: 0.906, right: 0.908).
} 
\revision{We note a lower Dice score for the amygdala in comparison to other brain structures across all methods. 
While the amygdala is a challenging structure to segment, also the small size can entail a lower Dice score.}

\begin{figure}[t]
\begin{center}
	\includegraphics[width=0.5\textwidth]{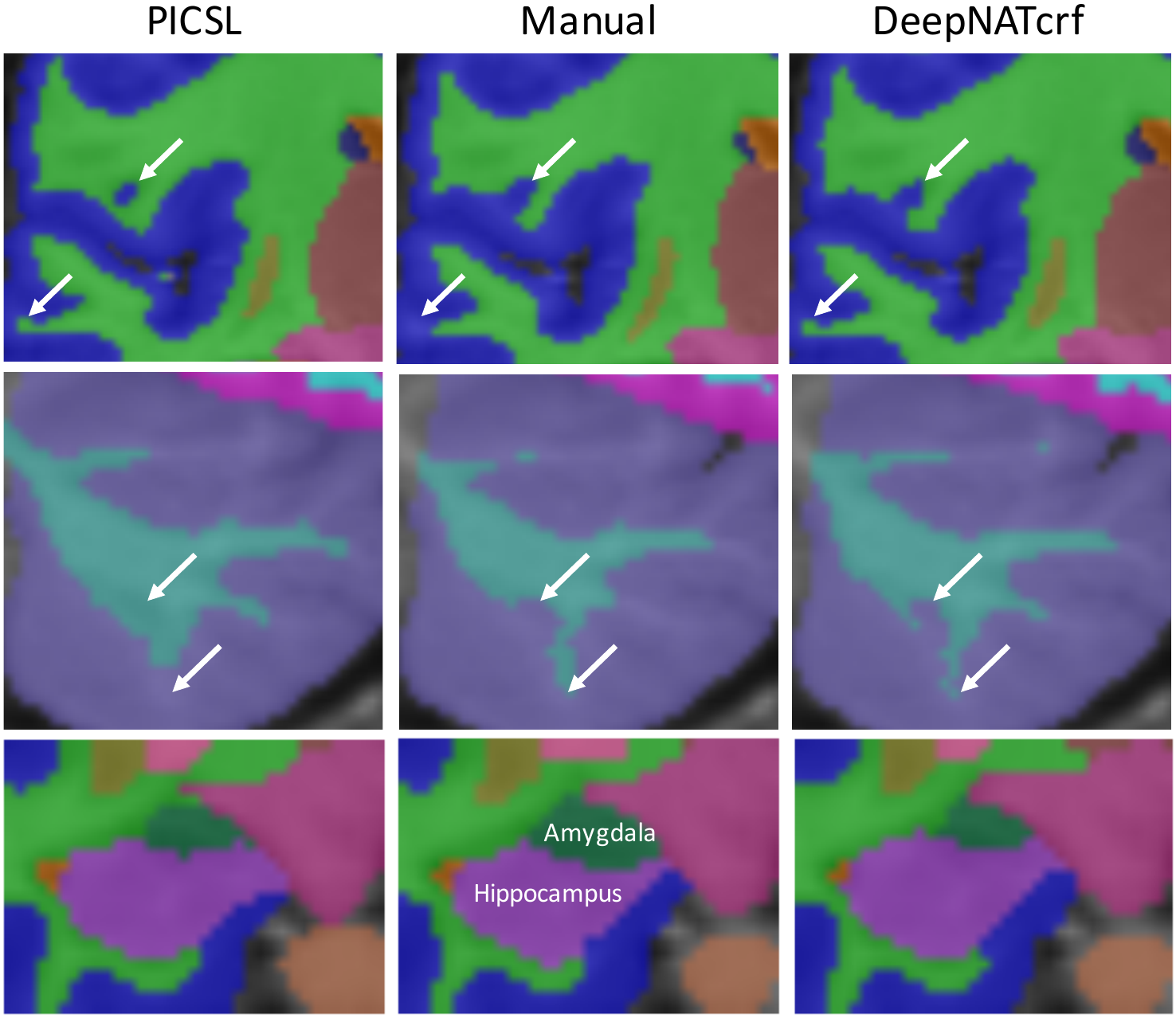} 
\caption{\revision{Example segmentations for PICSL and DeepNATcrf together with the corresponding manual segmentation. Overall, the segmentation quality of PICSL and DeepNATcrf is high, as already indicated by the quantitative results. 
\emph{First row}: Segmentation of the left cerebral white and gray matter region. 
DeepNATcrf better matches the manual segmentation in the center of the image, where PICSL produces an isolated region, see arrows. 
PICSL and DeepNATcrf produce an error at the bottom left part of the image, however the error is consistent among both approaches and the fine white matter region may have been skipped by the manual rater. 
\emph{Second row:} Segmentation of the right cerebellar white and gray matter. 
DeepNATcrf produces a more accurate segmentation of the gray matter region in the center of the image and better captures the thin white matter region at the bottom, as indicated by the arrows.
\emph{Third row:} Segmentation of the hippocampus and amygdala. PICSL produces a slightly more accurate segmentation of both structures. 
}
\label{fig:slicesZoom}
}
\end{center}
\end{figure}

\begin{figure}
\begin{center}
	\includegraphics[width=0.5\textwidth]{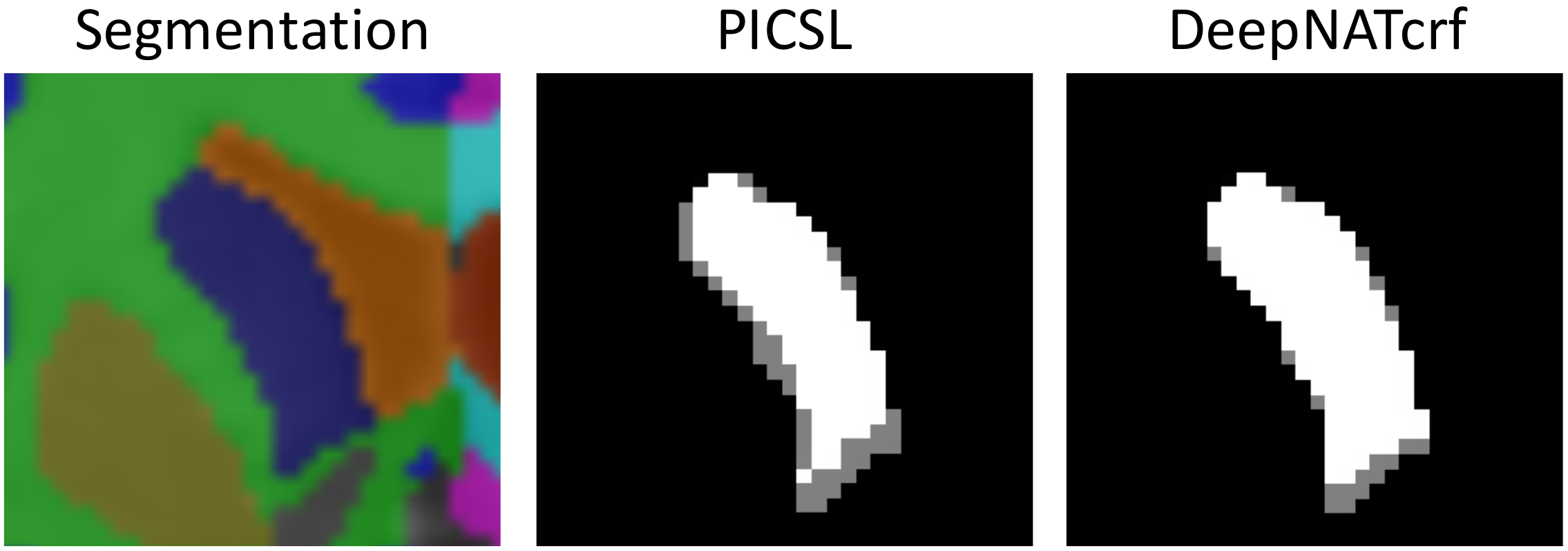} 
\caption{\revision{Example labeling of the left caudate, shown in dark blue on the segmentation mask.  
Segmentation masks on the middle and right panel show correct segmentations in white and segmentation errors in gray. 
DeepNATcrf shows a more accurate segmentation on the top left, center left, and bottom of the caudate. }
\label{fig:slicesZoomCaud}
}
\end{center}
\end{figure}

Figure~\ref{fig:slices} shows example segmentations for FreeSurfer, PICSL, and DeepNATcrf together with the manual segmentation. 
\revision{The results for PICSL and DeepNATcrf are very similar to the manual segmentation, while FreeSurer shows stronger variations, consistent with the quantitative results.} 
\revision{Figures~\ref{fig:slicesZoom} and~\ref{fig:slicesZoomCaud} illustrate zoomed in brain segmentations for structures with significant differences between PICSL and DeepNATcrf. 
In Figure~\ref{fig:slicesZoom}, segmentations of the cerebral white and gray matter as well as the cerebellar white and gray matter are more accurate with DeepNATcrf, whereas segmentations of the hippocampus and amygdala are more accurate with PICSL. 
Figure~\ref{fig:slicesZoomCaud} illustrates the segmentation of the caudate. 
The segmentation is illustrated by means of a segmentation map that highlights agreement and disagreement with the manual segmentation. 
Overall, DeepNATcrf is more consistent with the manual segmentation.}

\begin{figure*}[ht!]
\begin{center}
	\includegraphics[width=0.8\textwidth]{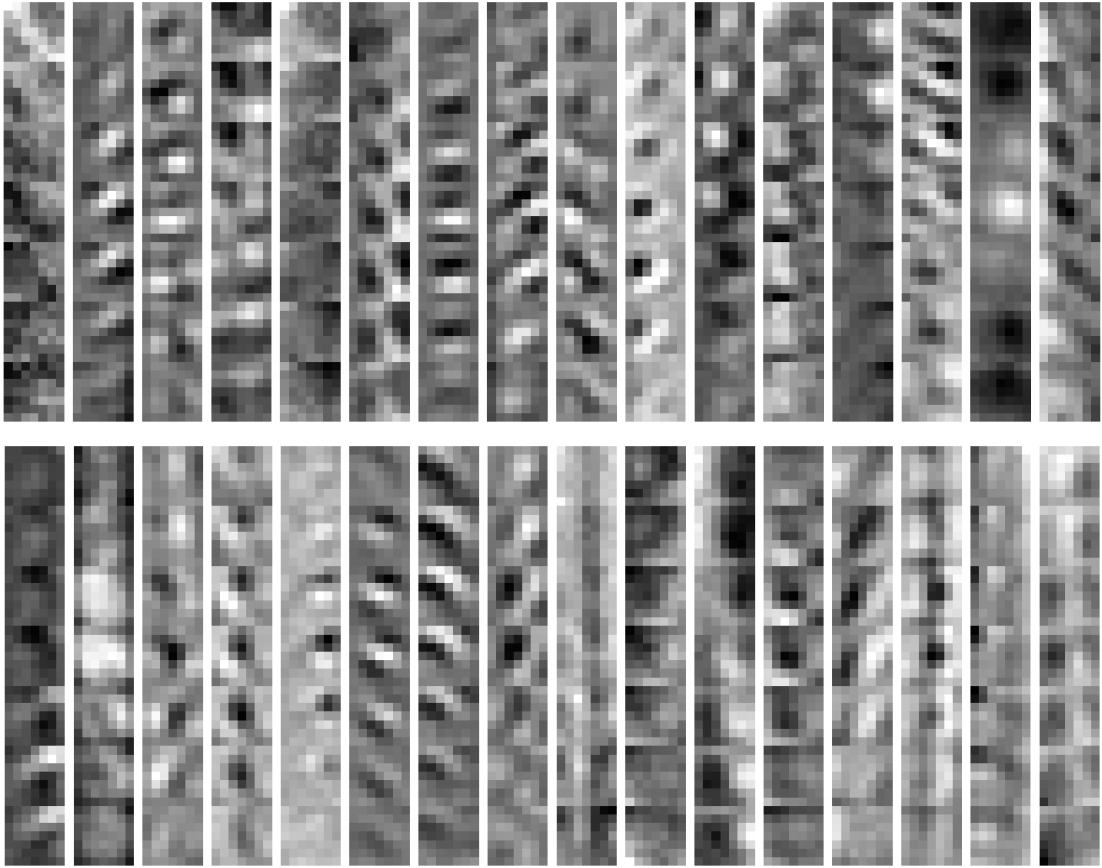} 
\caption{\revision{The figure shows 3D filters that were learned by the network in the first convolutional layer. The first conv layer consists of 32 filters of size $7 \times 7 \times 7$, as described in the architecture in Table~\ref{tab:archi}. We show all slices from the $7 \times 7 \times 7$ filter kernel vertically, so seven times a $7 \times 7$ patch. A number of these filter kernels resemble Gabor filters. The kernel patters further resemble 3D blob filters but also express more complex image patterns.}   \label{fig:filters}}
\end{center}
\end{figure*}

\revision{The convolutional layers in the DCNN can be interpreted as feature extractor from the image patch and the fully connected layers as classifier. 
To get a better understanding of the feature extraction, we show the learned convolutional filters of the first layer in Figure~\ref{fig:filters}.
The first layer consists of 32 filters of size $7 \times 7 \times 7$. 
The learned features are similar to 3D Gabor filters and 3D blobs. 
This is consistent with previous results on 2D DCNNs that report 2D Gabor filters and 2D color blobs on the first layer \citep{krizhevsky2012imagenet,yosinski2014transferable}. 
We do not include visualizations of filters from the second and third convolutional layers as they are less comprehensible due to the smaller filter size and the more abstract representation.
}

\begin{figure}[t]
\begin{center}
	\includegraphics[width=0.4\textwidth]{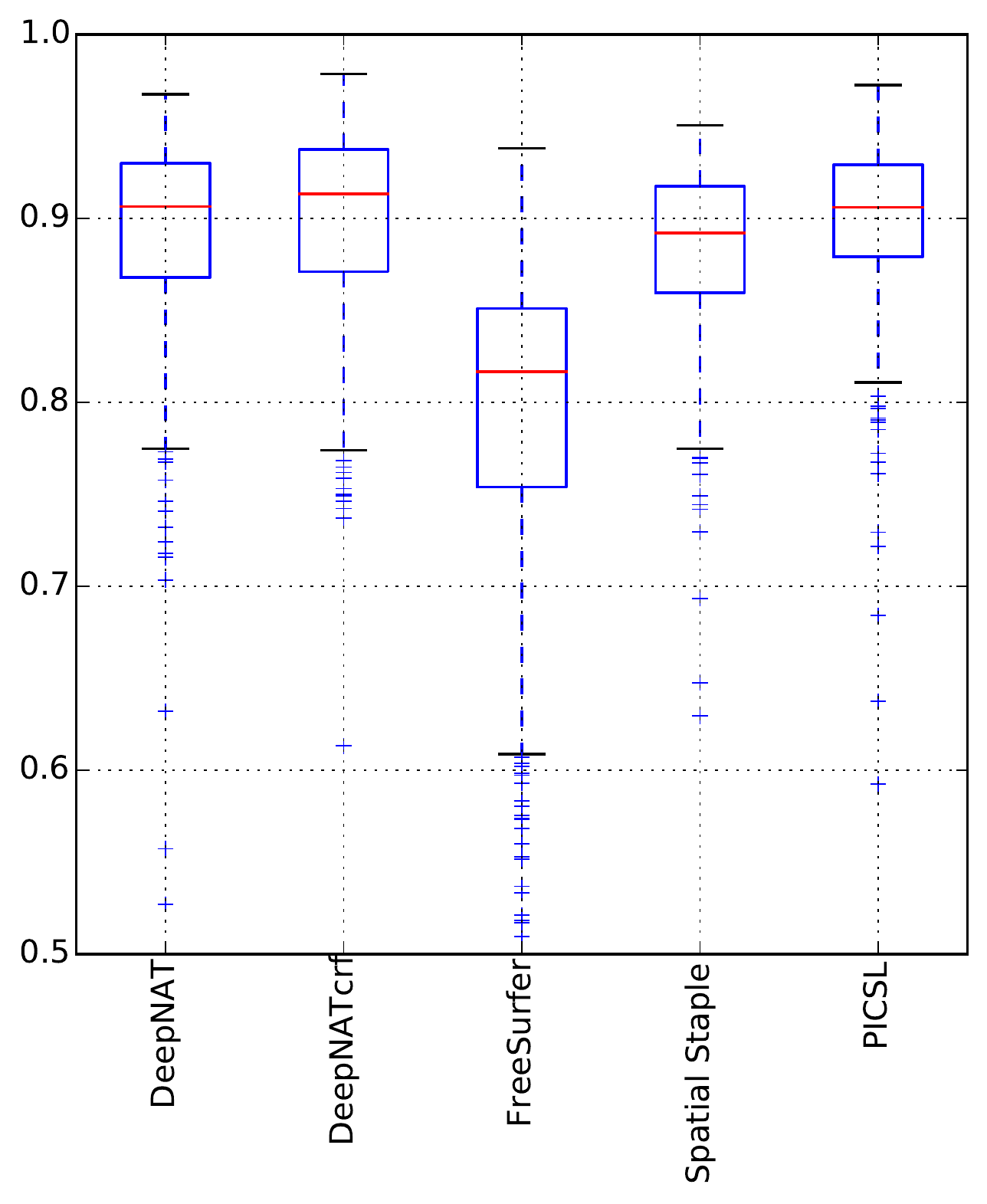} 
	\revision{
    \begin{tabular}{lr}
    Method & Median Dice \\ 
    \toprule
    DeepNAT &  0.906 \\
    DeepNATcrf &  \textbf{0.913}  \\
    FreeSurfer & 0.817 \\ 
    Spatial STAPLE & 0.892 \\ 
    PICSL & 0.906 \\
  \end{tabular} }	  
\caption{\revision{Segmentation results similar to Fig.~\ref{fig:boxComp} for decreasing the training set in DeepNAT from 20 to 15 images and increasing the testing data for all methods from 10 to 15.  %
In the figure, red lines indicate the median, the boxes extend to the $25^{\text{\tiny th}}$ and $75^{\text{\tiny th}}$ percentiles, and the whiskers reach to the most extreme values not considered outliers (crosses).  
The table lists the median Dice for the different approaches. }
\label{fig:boxComp15}
}
\end{center}
\end{figure}

\begin{figure}[t]
\begin{center}
	\includegraphics[width=0.4\textwidth]{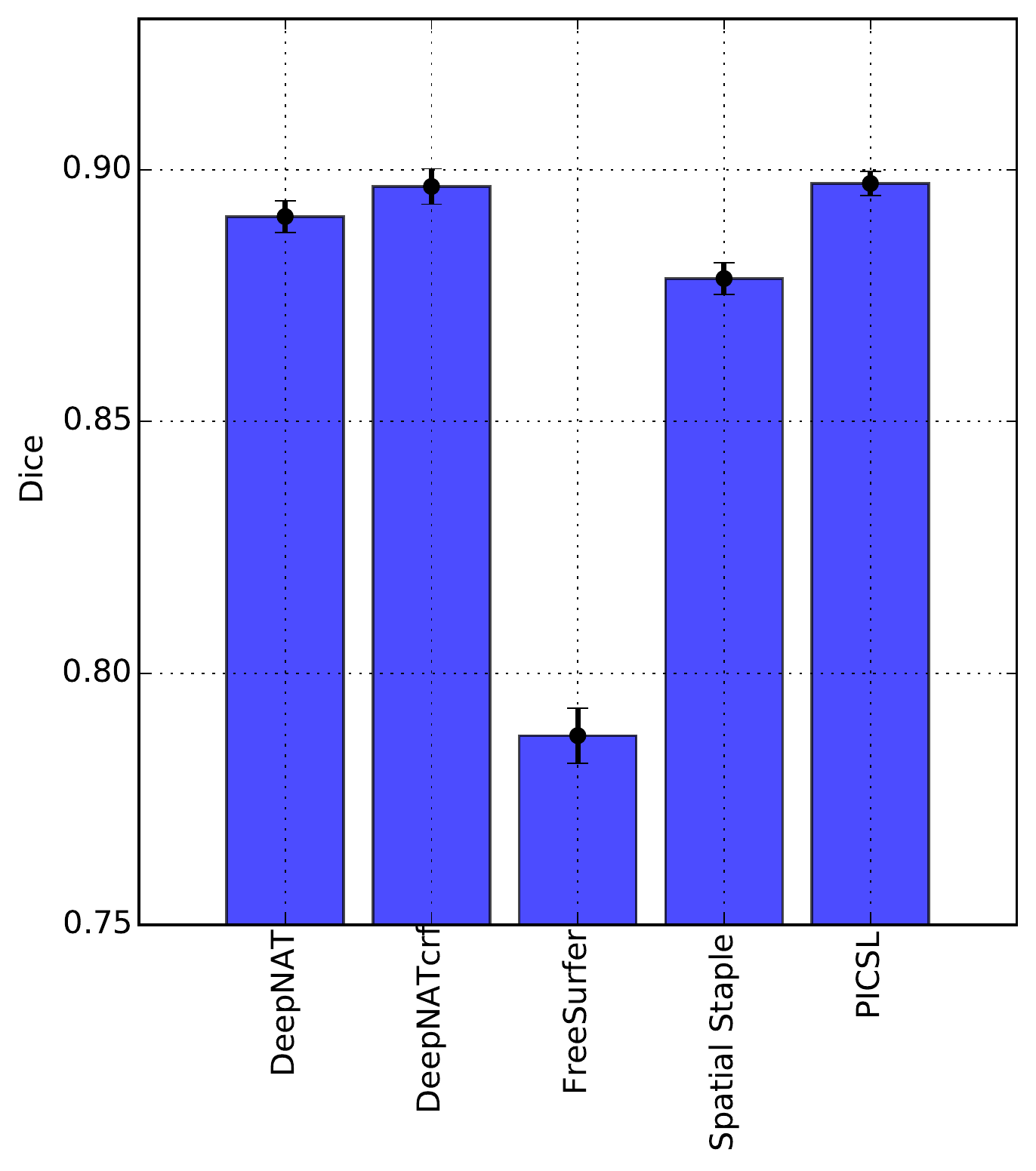} 
	\revision{
    \begin{tabular}{lr}
    Method & Mean Dice \\ 
    \toprule
    DeepNAT &  0.891 \\
    DeepNATcrf &  \textbf{0.897}  \\
    FreeSurfer & 0.788 \\ 
    Spatial STAPLE & 0.878 \\ 
    PICSL & \textbf{0.897} \\
  \end{tabular}
  }	  
\caption{\revision{Segmentation results in Dice for DeepNAT and DeepNATcrf  together with alternative segmentation methods: FreeSurfer, spatial STAPLE, and PICSL. 
The bars show the mean Dice score and the lines show the standard error. 
The table lists the mean Dice for the different approaches. } 
\label{fig:boxCompMean15}
}
\end{center}
\end{figure}

\revision{Finally, we reduce the training set from 20 to 15 and increase the testing set from 10 to 15 to have the identical setup to the labeling challenge. 
We employ data augmentation with jittering to counter the reduction in training data and increase the training time to 50 epochs. 
Figures~\ref{fig:boxComp15} shows the results over all 25 brain structures with the median and percentiles and Figure~\ref{fig:boxCompMean15} shows the mean and standard error.
We note a slight overall decrease in accuracy across all methods, compared to Figures~\ref{fig:boxComp} and~\ref{fig:boxCompMean}, as a result of modifying the testing data. 
For 15 training and 15 test images, DeepNATcrf yields significantly higher Dice scores in comparison to DeepNAT ($p < 0.001$), FreeSurfer ($p < 0.001$), and STAPLE ($p < 0.001$), whereas the difference to PICSL ($p=0.06$) is not significant.
The median of DeepNATcrf is 0.007 Dice points higher than PICSL, whereas the mean Dice points are the same. 
The decreasing gap between DeepNATcrf and PICSL in testing accuracy is likely associated with the the reduction of the training set for learning the network. 
}

\section{Discussion}

\textbf{DeepNAT architecture:} 
One of the biggest challenges when working with deep convolutional neural networks is the vast number of decisions to take for the specification of the architecture. 
Many of the decisions are a trade-off between additional discriminative power of the network and training complexity as well as memory requirements. 
For instance, we do not use a batch normalization after the first convolution to avoid the high memory consumption. 
An alternative design for the convolutional stage would have been to work with smaller kernels of size~3 and to build a deeper hierarchy, similar to VGG~\citep{simonyan2014very}. 
We have not fully explored this direction, also due to long training times, but initial results did not look very promising.

In this work, we used 3D convolutional neural networks for brain segmentation. 3D DCNNs have been used for medical applications before~\citep{li2014deep,brebisson2015deep}, however, the majority of work is on 2D or 2.5D applications. 
Given that we deal with the segmentation of 3D MRI scans, it  seems natural to work with a 3D network for the classification. 
Yet, working with a 3D network yields an increase in complexity because the convolutional filters and the internal representations have an additional dimension. %
By employing batch normalization, dropout, and the Xavier initialization, we are able to train 3D networks with more layers than previous 3D DCNNs%
, where deeper networks can model more complex relationships between input and output data. %

In many image segmentation tasks, we are facing the challenge of dealing with a large background class that surrounds the structures of interest. 
The background typically consists of multiple structures that are of no further interest to the application and merged into the background class. 
For multi-atlas segmentation, we have reported that the dominant background class can cause an under-segmentation of the target structure, because it introduces a bias in the label estimation~\citep{wachinger2014atlas}.  
Here, we address the class imbalance problem with a hierarchical approach by first separating foreground from background and then identifying the individual brain structures on the foreground. %
Our results show the benefit of this cascaded approach in comparison to directly segmenting brain structures.

\textbf{Location information:}
A drawback of patch-based segmentation methods is the loss of the larger image context, given that brain scans from different subjects are overall fairly similar. %
Context information can be crucial for differentiating small image regions across the brain that can appear very similar due to symmetries. 
To retain context information, we include location information in the network. 
The results demonstrate that the addition of coordinates leads to a substantial increase in segmentation accuracy. 
In this work, we introduced spectral brain coordinates, a parameterization of the brain solid with Laplace eigenfunctions, which yielded an improvement over Cartesian coordinates.  
Interestingly, the combination of spectral and Cartesian coordinates resulted in a further increase in segmentation accuracy, indicating that they contain complementary information.

\textbf{Multi-task learning:}
Multi-task learning has several applications in machine learning, but we have not yet seen its application for image segmentation. 
Instead of only predicting the label of the center voxel, we simultaneously learn and predict also the labels of the neighboring voxels. 
Our results show that multi-task learning yields a significant improvement over single-task segmentation for all brain structures. 
This is consistent with results from non-local means segmentation, where results from the multi-point method showed improvements over the single-point approach~\citep{rousseauHS11}. 
Multi-task learning leads to several predictions per voxel, which can generate more robust segmentations by overruling incorrect predictions. 
The tasks are learned by sharing the same network, with only the last layer specializing on a single task. 
This causes only a small increase in the overall number of parameters. 
We have experienced a faster convergence of the multi-task network compared to the single-task network, which may be attributed to the enforcement of promising gradient directions from all simultaneous tasks. 
This is consistent with previous observations from multi-task learning for sequence to sequence modeling~\citep{luong2015multi}.

We observe that the center task has a slightly but consistently higher accuracy than the surrounding tasks. 
This is surprising because no priority or higher weighting was assigned to the center task. 
One possible explanation could be that the center location has a larger context but when considering that a patch size of 23 was used, this should not have a strong impact. 
It rather seems that the convolutional stage of the network with convolution filters and max-pooling better captures the information for predicting the center label.

\textbf{Comparison to state-of-the-art:}
In our results, we compare to FreeSurfer and two methods from the MICCAI labeling challenge, PICSL and spatial STAPLE. 
FreeSurfer is one of the most commonly used tools for brain anatomy reconstruction in practice. 
It performed worse than the other methods in the comparison, however, all other methods used the provided training dataset, whereas FreeSurfer uses its own atlas. 
Dataset bias may therefore play a role. 
In addition, the protocol for the manual labeling of the scans may not be entirely consistent. 
PICSL was the winner of the segmentation challenge and spatial STAPLE was among the best performing methods. 
Both of these approaches are based on a multi-atlas approach, where all atlas images are registered to the test image. 
\revision{A single registration takes about 2 hours of runtime, so that the registration of all 15 training images takes about 30 hours.} 
The registration can be time-consuming for many image pairs, consequently scaling such methods to larger atlases seems challenging. 
In contrast, the inclusion of additional training data does not affect testing time for DeepNAT, which is about 1 hour. 
We trained the final DeepNAT model for about three days on the GPU, but also PICSL is based on an extensive training of the corrective classifier, which was reported with 330 CPU hours. 
\revision{The runtime of DeepNAT could be further improved by using cuDNN and accounting for overlapping patches.}

The results of DeepNAT resulted in statistically significant improvements over FreeSurfer and spatial STAPLE. 
DeepNAT in combination with the CRF yielded the overall highest median Dice score, but the improvement over PICSL is not statistically significant. 
\revision{Performing tests on the per structure level resulted in advantages for DeepNAT for cortical structures, which may be explained by the difficulty in registering complex folding patterns.} 
For subcortical structures, the results were not as clear. 
\revision{The variation in significance for the left and right caudate is driven by varying results of PICSL, but the source of the difference is not clear as no preference to one of the hemispheres seems to be given in PICSL.}

\textbf{Conditional Random Field:}
Our results demonstrate the benefit of inferring the final, discrete segmentation from the probabilistic network outcome with the fully connected conditional random field. 
Previous applications of the fully connected CRF have been for 2D applications. %
The pairwise constraints formulated in the CRF ensure label agreement between close voxels. 
In the appearance term of the pairwise potential, we use the difference of voxel intensities as a measure of similarity. 
Such similarity terms have been extensively studied in spectral clustering for image segmentation~\citep{shi2000normalized}, where the concept of the intervening contour was proposed~\citep{fowlkes2003learning} and adapted for medical image segmentation~\citep{wachinger2015contour}. 
Integrating the concept of intervening contours into the pairwise potentials of the CRF seems promising to further improve segmentation accuracy. 
\revision{Note that we do not train the CRF, so while DeepNAT is an end-to-end learning system, DeepNATcrf is not.}

\textbf{Training Data:}
One of the big issues when using deep learning in the \revision{medical} domain is the access to a large enough training dataset. 
\revision{The training set used in our experiments seems small for training a deep convolutional neural network with millions of parameters compared to the millions of images from ImageNet typically used in computer vision.}
However, DeepNAT does not directly predict the segmentation of the entire image but only of image patches. 
Working with patches makes the training feasible as each scan contains millions of patches that can be extracted for learning. 
In the future, it would be interesting to further explore ideas about directly estimating the segmentation of the entire image without the reduction to patches. 
This can lead to a drastic speed-up, due to the computational overhead when working with overlapping patches. 
Yet, such an approach would require a much larger number of images with manual segmentations for training, which are very time consuming to create. 

\revision{Due to the limited size of the dataset, we have not split between validation and testing set. 
We have directly compared the different contributions in DeepNAT (coordinates, hierarchy, multi-task) on the testing set, see Figure~\ref{fig:boxMethod}.
Consequently, there is a risk of overfitting on the testing data. 
However, these comparisons involved conceptual design decisions and not a detailed parameter fine-tuning, so we consider the risk of overfitting to be limited. 
Further, the good performance of DeepNAT persisted after reducing the training dataset to 15 scans and increasing the testing dataset to 15 scans.}

\revision{%
DeepNAT may be specifically adapted for segmenting young, old, or diseased brains by fine-tuning. 
The large potential of fine-tuning pre-trained models for deep learning has been shown previously. 
In the medical imaging domain, \cite{gao2016} fine-tuned weights trained on ImageNet to detect lung disease in CT images. 
\cite{yosinski2014transferable} show that transferability of features, e.g., convnets trained on ImageNet and then fine-tuned to other tasks, depends on how general those features are; the transferability gap increases as the distance between tasks increases and features become less general.
Notably, these studies operate on 2D images and we are not aware of work that fine-tunes networks with volumetric input, where the pre-trained models of DeepNAT can provide a first step in this direction.}

\section{Conclusion}
We presented DeepNAT, a 3D deep convolutional neural network for brain segmentation of structural MRI scans. 
The main contributions were (i)~multi-task learning, (ii)~hierarchical segmentation, (iii)~spectral coordinates, and (iv)~a 3D fully connected conditional random field. 
Multi-task learning simultaneously learns the label prediction in a small neighborhood. 
Spectral coordinates form an intrinsic parameterization of the brain volume and provide context information to patches. 
The hierarchical approach accounts for the class imbalance between the background class and separate brain structures. 
And finally, the conditional random field ensures label agreement between close voxels. 
We train the 3D network by integrating latest advances in deep learning to initialize weights, to correct for internal covariate shift, and to limit overfitting for training such complex models. 
Our results demonstrated the high potential of convolutional neural networks for segmenting neuroanatomy. %

All in all, image segmentation is a well-suited task for convolutional neural nets, which are arguably at the forefront of the the deep learning wave. 
The segmentation accuracy of convolutional neural nets is likely to further improve in the future, given the increasing amount of training data, methodological advances for deep networks, and progress in GPU hardware. 
We believe that the purely learning-based approach with neural networks offers unique opportunities for tailoring segmentations to young, old, or diseased brains. 
While it may be difficult to obtain enough training data on such specific applications, fine-tuning a pre-trained network seems like a promising avenue.

\revision{Our extensions to caffe, network definitions and trained networks are available for download: \\ \url{https://tjklein.github.io/DeepNAT/}.}

\section{Acknowledgement}
Support for this research was provided in part by the Humboldt foundation, SAP SE, Förderprogramm für Forschung und Lehre, the Bavarian State Ministry of Education, Science and the Arts in the framework of the Centre Digitisation.Bavaria (ZD.B), the National Cancer Institute (1K25CA181632-01), the Massachusetts Alzheimer's Disease Research Center (5P50AG005134), the MGH Neurology Clinical Trials Unit, the Harvard NeuroDiscovery Center, Genentech (G-40819), and the NVIDIA Corporation.

\section{References}

\bibliographystyle{elsarticle-harv}
\bibliography{jab_bib}

\end{document}